\documentclass{article}
\usepackage{spconf,amsmath,graphicx,hyperref}
\usepackage{multirow}
\usepackage{fancyvrb}
\usepackage{xcolor}
\usepackage{listings}
\usepackage{booktabs}
\usepackage{arydshln}

\title{Up to 36x Speedup: Mask-based Parallel Inference Paradigm for Key Information Extraction in MLLMs}
\name{Xinzhong Wang$^{1\dagger}$, Ya Guo$^{2\dagger}$, Jing Li$^{2}$, Huan Chen$^{2}$,  Yi Tu$^{2}$, Yijie Hong$^{1}$, Gongshen Liu$^{13\ddagger}$, Huijia Zhu$^{2\ddagger}$
      \thanks{$^{\dagger}$ Equal contribution.}
      \thanks{$^{\ddagger}$ Corresponding authors.}
}
\address{
  $^1$Shanghai Jiao Tong University\\
  $^2$Ant Info Security Lab, Ant Group\\
  $^3$Inner Mongolia Research Institute, Shanghai Jiao Tong University, Hohhot 010010
}
\begin{document}
\ninept
\maketitle
\begin{abstract}
Key Information Extraction (KIE) from visually-rich documents (VrDs) is a critical task, for which recent Large Language Models (LLMs) and Multi-Modal Large Language Models (MLLMs) have demonstrated strong potential. However, their reliance on autoregressive inference, which generates outputs sequentially, creates a significant efficiency bottleneck, especially as KIE tasks often involve extracting multiple, semantically independent fields. To overcome this limitation, we introduce \textbf{PIP}: a \textbf{P}arallel \textbf{I}nference \textbf{P}aradigm for KIE. Our approach reformulates the problem by using ``[mask]'' tokens as placeholders for all target values, enabling their simultaneous generation in a single forward pass. To facilitate this paradigm, we develop a tailored mask pre-training strategy and construct large-scale supervised datasets. Experimental results show that our PIP-models achieve a \textbf{5–36×} inference speedup with negligible performance degradation compared to traditional autoregressive base models. By substantially improving efficiency while maintaining high accuracy, PIP paves the way for scalable and practical real-world KIE solutions. 
\end{abstract}

\section{Introduction}

Key Information Extraction (KIE) aims to extract and structure key information (e.g., names, dates, amounts) from visually-rich documents (VrDs) like invoices and forms. As a downstream task of document understanding, it requires integrating multimodal features, including text, layout, and visual cues. Recent advances in Large Language Models (LLMs) \cite{llama,llama2,vicuna,glm45,textmonkey,claude} and particularly Multi-Modal Large Language Models (MLLMs) \cite{gpt4v,deepseek-vl,deepseekocr,gemini,gemini1.5,minigpt,llavanext} have shown remarkable performance on KIE tasks, reshaping the field.

\begin{figure}[ht]
\centering 
\includegraphics[width = \linewidth]{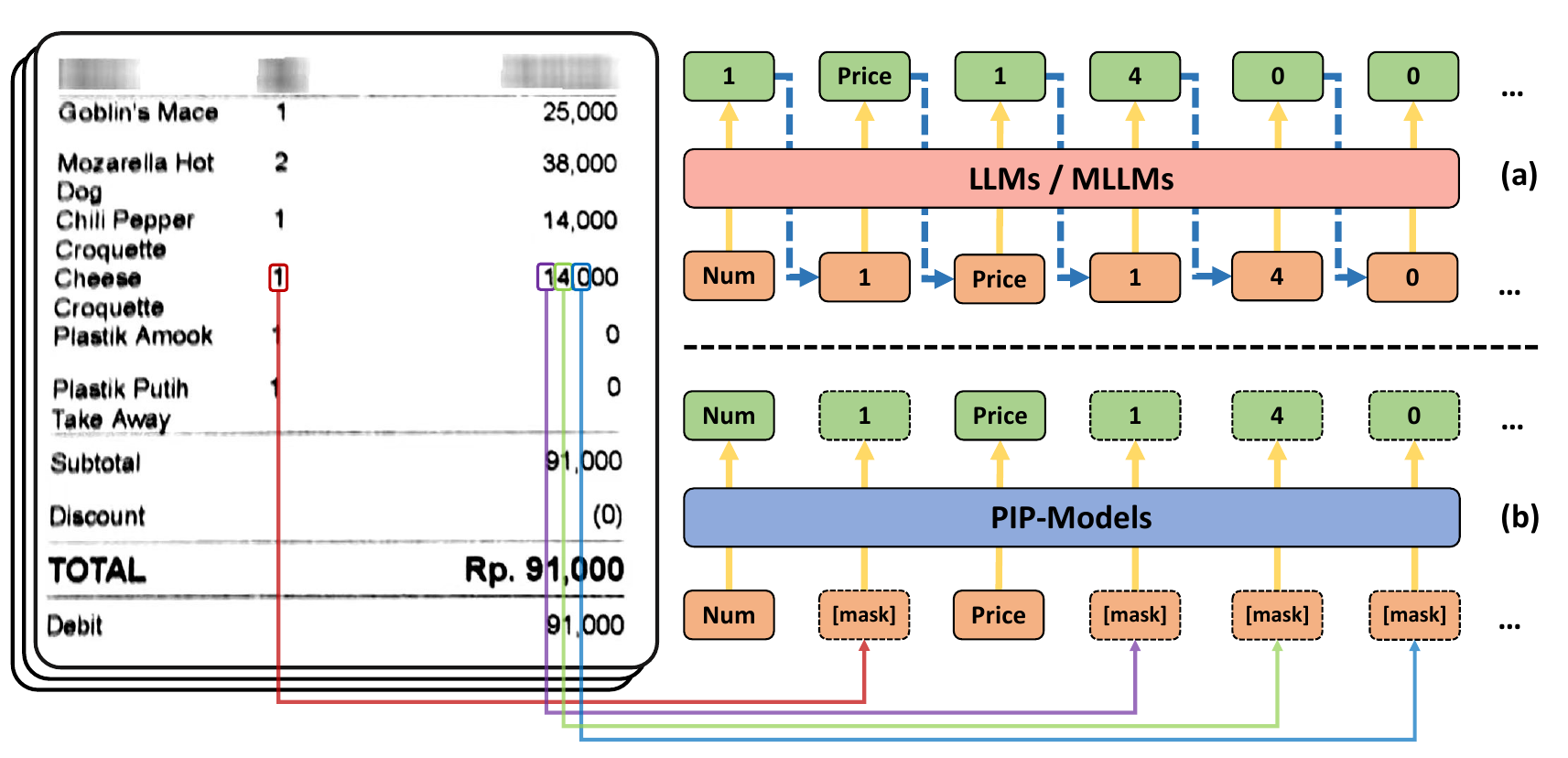}
\caption{Comparison of two inference paradigms: (a) Traditional autoregressive inference, which generates tokens sequentially one by one; (b) Our PIP-Models, where different ``[mask]'' tokens independently attend to distinct image regions and generate all tokens in parallel.}
\label{fig:parallel_reason}
\end{figure}

\begin{figure*}[ht]
\centering 
\includegraphics[width=0.8\linewidth]{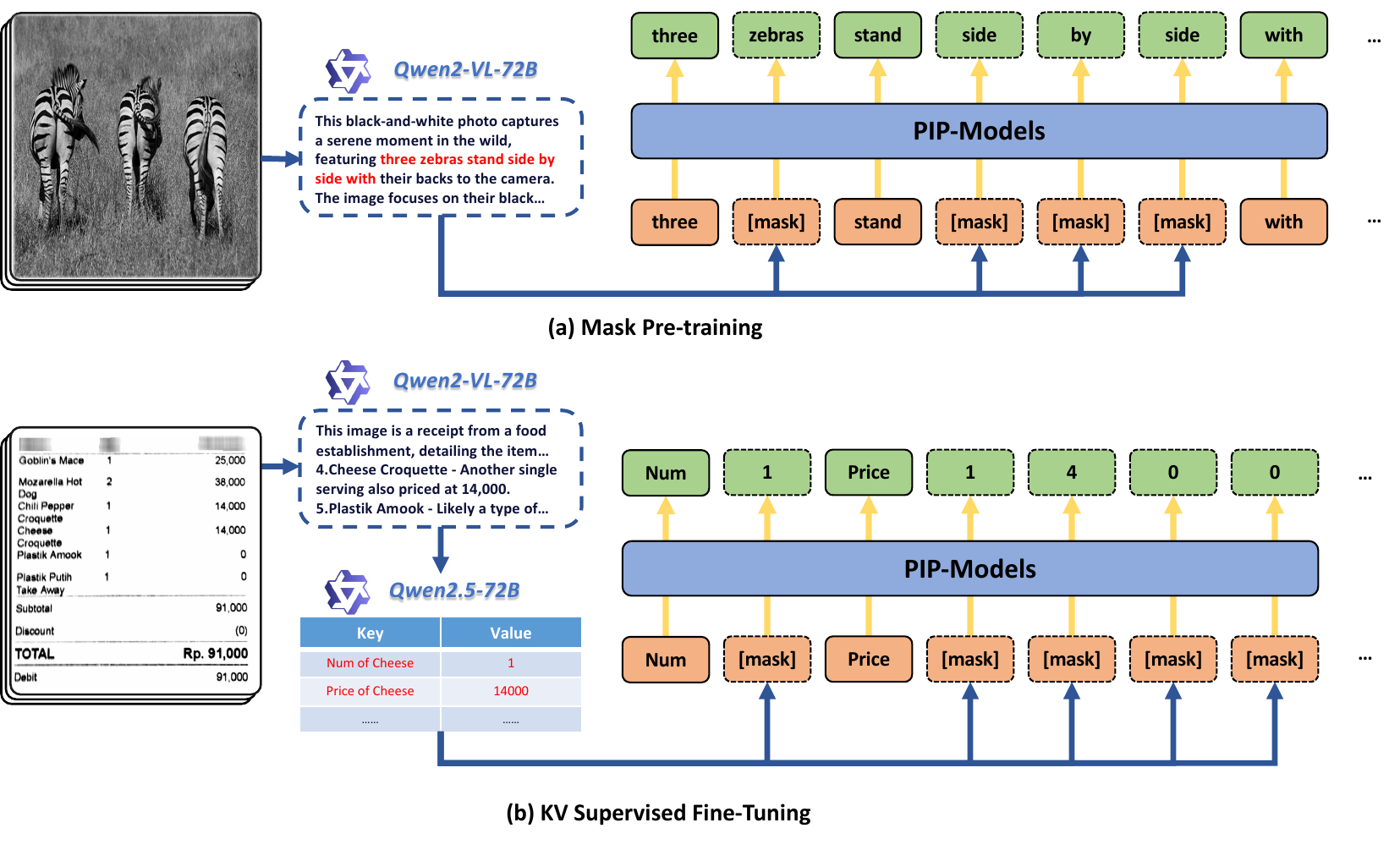}
\caption{Our overall training processes: (a) represents the mask pre-training phase, and (b) denotes the KV supervised fine-tuning stage.}
\label{fig:training_process}
\end{figure*}

While LLMs require a preliminary OCR step to process VrDs, this two-stage approach suffers from error propagation and high computational overhead. MLLMs mitigate these issues by processing images and text in an end-to-end fashion, making them a more promising foundation for KIE. However, both LLMs- and MLLMs-based methods are constrained by the autoregressive inference paradigm, which generates tokens sequentially.

This sequential generation is suboptimal for KIE. The extraction of distinct fields—such as the ``Num'' and ``Price'' for an item in Figure \ref{fig:parallel_reason}—are often semantically independent sub-tasks that are inherently parallelizable. We argue that even tokens within a single answer can be generated in parallel. As KIE is largely a retrieval task, each output token (e.g., ``14000'') corresponds to a specific visual region. Its generation thus depends more on attending to the correct image location than on previously generated tokens.

To address this limitation, we propose \textbf{PIP}: a simple yet effective \textbf{P}arallel \textbf{I}nference \textbf{P}aradigm for KIE. PIP reformulates the task by replacing target values in the prompt with ``[mask]'' tokens (e.g., ``Num:[mask][mask]... Price:[mask][mask]...''). This formulation allows the model to decode all masked positions in parallel within a single forward pass, dramatically reducing inference latency.

While mask-based parallel decoding has been explored in unimodal contexts \cite{padellm}, its application to multimodal KIE is non-trivial due to potential interference between masked tokens processing complex visual and textual inputs. Our key insight is that the spatial correspondence between KIE outputs and document regions allows each ``[mask]'' token to focus on a distinct image area, naturally minimizing interference. We validate this hypothesis through attention visualization in later section, establishing the feasibility of parallel decoding for multimodal KIE.

To adapt MLLMs to this paradigm, we introduce a dedicated mask pre-training stage and construct a large-scale supervised fine-tuning dataset of key-value (KV) pairs. Experiments show that our approach achieves a \textbf{5–36×} improvement in inference speed with negligible performance degradation compared to the autoregressive baselines.

Our main contributions are:

\begin{itemize}
\item We introduce PIP, a parallel inference paradigm that reformulates KIE for significant efficiency gains.
\item We develop a specialized training methodology, including a mask pre-training stage and a large-scale supervised fine-tuning dataset, to enable MLLMs to perform parallel decoding.
\item Extensive experiments demonstrate that our method accelerates KIE inference by 5–36× while maintaining comparable performance to autoregressive models.
\end{itemize}

\section{Methodology}

\subsection{Task Formulation}

Traditional MLLMs for KIE employ an autoregressive paradigm, generating tokens sequentially. Given an image $I$, they maximize the joint probability of a text sequence $X = [x_1, \dots, x_T]$:

\begin{equation}
P(x_1, \dots, x_T \mid I) = \prod_{t=1}^{T} P(x_t \mid x_1, \dots, x_{t-1}, I).
\end{equation}

This word-by-word generation process, where each token $x_t$ is predicted based on preceding tokens, is inherently sequential and slow, limiting real-world applicability.

\begin{equation}
x_t = \arg\max_{x} P(x \mid x_1, \dots, x_{t-1}, I).
\end{equation}

To address this bottleneck, we reformulate the task into a parallel inference paradigm. The model is given an input sequence $X$ with a set of masked positions $M \subseteq \{1, \dots, T\}$. Its objective is to predict all masked tokens $x_M$ simultaneously based on the unmasked tokens $x_{\setminus M}$ and the image $I$:

\begin{equation}
P(x_M \mid x_{\setminus M}, I) = \prod_{t \in M} P(x_t \mid x_{\setminus M}, I).
\end{equation}

This allows the model to predict all missing tokens in a single forward pass, eliminating the sequential dependency of autoregressive models and significantly improving inference efficiency.

\begin{equation}
\hat{x}_t = \arg\max_{x} P(x \mid x_{\setminus M}, I) \quad \forall t \in M.
\end{equation}

\begin{table*}
\centering
\resizebox{0.6\linewidth}{!}{
\renewcommand{\arraystretch}{1.0}
\begin{tabular}{clcccccccc}
\toprule
     & & \multicolumn{2}{c}{\textbf{FUNSD}} &  & \multicolumn{2}{c}{\textbf{SROIE}} & & \multicolumn{2}{c}{\textbf{CORD}} \\
\cline{3-4} \cline{6-7} \cline{9-10}
& \textbf{Models} & ANLS & Time(s) & & ANLS & Time(s) & & ANLS & Time(s) \\
\hline
\multirow{4}{*}{\textbf{LLMs}}
& Llama2-7B &$40.8^*$ & 0.692 & &$4.4^*$ & 0.753 & & $15.9^*$ & 0.237 \\
& Vicuna-1.5-7B &$48.1^*$ &0.779 & & $51.4^*$ & 0.861 & & $68.2^*$ & 0.249 \\
& LayoutLLM-7B & $80.0^*$ & - & & $63.1^*$ & - & & $72.1^*$ & - \\
& LayTextLLM-7B & $\textbf{81.0}^\sim$ & 1.055 & & $96.1^\sim$ & 0.948 & & $82.5^\sim$ & 0.339 \\
\hline
\multirow{4}{*}{\textbf{MLLMs}}
& LLaVAR-7B & $1.7^*$ &0.327 &  & $13.6^*$ & 0.289 & & $2.4^*$ & 0.218 \\
& LLaVA-1.5-7B & $1.9^*$ & 0.310 & & $18.1^*$ & 0.269 & & $3.8^*$ & 0.207 \\
& InternVL2-8B & 61.2 & 0.615 & & 95.1 & 0.541 & & 88.2 & 0.314 \\
& Qwen2-VL-7B & 77.9 & 0.455 & & 94.1 & 0.532 & & 91.2 & 0.288 \\
\hline
\multirow{2}{*}{\textbf{Ours}}
& PIP-InternVL2-8B & 72.3 & 0.064 & & 93.4 & \underline{0.034} & & 93.1 & \underline{0.022} \\
& PIP-Qwen2-VL-7B & 79.3 & \underline{0.053} & & \textbf{97.0} & 0.051 & & \textbf{97.3} & 0.028 \\
\bottomrule 
\end{tabular}}
\caption{Results of different models for KIE tasks. * denotes the results from \cite{layoutllm}, $\sim$ indicates that the results are from \cite{laytextllm}. The best results are highlighted in \textbf{bold}, while the shortest inference times are \underline{underscored}.}
\label{tab:KIE_results}
\end{table*}

\subsection{Model Training Process}

Our PIP-Models are built on existing autoregressive MLLMs. To enable parallel inference, we introduce a two-stage training process: \textbf{Mask Pre-training} and \textbf{KV Supervised Fine-Tuning}, as illustrated in Figure \ref{fig:training_process}. This process transitions the model from sequential to parallel generation, boosting inference speed for KIE tasks without sacrificing accuracy.

\subsubsection{Mask Pre-training}
This stage adapts the base MLLMs to our parallel framework using a mask-and-predict scheme on large-scale image-caption data. For each sample, we randomly mask a fraction of the caption tokens. The model is trained to reconstruct the masked tokens given the visible tokens and the image, forcing it to learn non-sequential text generation.

Pivotal to this stage is replacing the model's unidirectional causal attention with a bidirectional attention mechanism, following LLaDA \cite{llada}. Unlike autoregressive models where tokens only attend to predecessors, bidirectional attention allows each token to attend to all other tokens in the sequence. This provides a complete context for each prediction, which is crucial for mitigating error propagation and essential for understanding the global document layout in KIE tasks.

\textbf{Data:} The pre-training dataset comprises 13 million images with detailed captions generated by Qwen2-VL-72B \cite{qwen2vl}. It covers a diverse range of content including documents, landscapes, and artworks, providing a broad foundation for learning robust parallel inference capabilities.

\subsubsection{KV Supervised Fine-Tuning}
While mask pre-training imparts parallel inference capabilities, the model requires task-specific adaptation for KIE. This supervised fine-tuning stage uses a curated KV extraction dataset to refine the model's ability to identify and extract structured information from documents, enhancing accuracy and reducing content hallucination.

\textbf{Data:}
The fine-tuning dataset was meticulously curated. We prioritized high-resolution images, filtered out samples with poor OCR quality, categorized documents into 48 classes, and anonymized all personally identifiable information.

Our annotation pipeline began with pre-annotation using MLLMs (Figure \ref{fig:training_process}b). Qwen2-VL 72B \cite{qwen2vl} generated descriptive captions, which were then parsed by Qwen2.5 72B \cite{qwen25} to extract structured KV pairs. To mitigate hallucination, the model was trained to output "unknown" for keys not present in the image. Subsequently, all machine-generated annotations underwent a human-in-the-loop verification process to correct inaccuracies and ensure dataset fidelity.


\section{Experiment}

\subsection{Experiments Setup}
\textbf{Datasets:}
We evaluate our method on five public benchmarks for KIE: \textbf{FUNSD} \cite{funsd}, \textbf{SROIE} \cite{sroie}, \textbf{CORD} \cite{cord}, \textbf{POIE} \cite{poie}, and \textbf{WildReceipt} \cite{wildreceipt}.

\noindent \textbf{Baselines:}
We compare against OCR-based LLMs (\textbf{Llama2-7B} \cite{llama2}, \textbf{Vicuna-1.5-7B} \cite{vicuna}, \textbf{LayoutLLM} \cite{layoutllm}, \textbf{LayTextLLM} \cite{laytextllm}) and OCR-free MLLMs (\textbf{LLaVAR-7B} \cite{llavar}, \textbf{LLaVA-1.5-7B} \cite{llava}). Our approach is built upon \textbf{InternVL2-8B} \cite{internvl2} and \textbf{Qwen2-VL-7B} \cite{qwen2vl}, which also serve as baselines.

\noindent \textbf{Evaluation Metrics: }
Performance is assessed by Average Normalized Levenshtein Similarity (ANLS) for \textbf{FUNSD}, \textbf{SROIE}, and \textbf{CORD}, and F1 score for \textbf{POIE} and \textbf{WildReceipt}. Inference efficiency is measured as the average time per sample on 8×A100 GPUs.

\subsection{Results}
We present a comparative analysis of our proposed framework, benchmarking it against state-of-the-art (SOTA) methods and the original base models across multiple dimensions.

\subsubsection{Compared with SOTA Models}

Table \ref{tab:KIE_results} provides a comprehensive evaluation of our method against leading LLMs and MLLMs, focusing on extraction performance and inference efficiency.

\textbf{Extraction Performance:}
The FUNSD dataset is particularly challenging due to its dynamic key-value schema, leading to lower ANLS scores compared to datasets with fixed structures like SROIE and CORD. General-purpose LLMs (e.g., Llama2-7B, Vicuna-1.5-7B) are limited by the generic text-generation paradigm, reaching a maximum ANLS of 69.0. Layout-aware models like LayTextLLM significantly improve upon this by incorporating spatial features, setting a new SOTA on FUNSD with an ANLS of 81.0. Vision-based MLLMs (e.g., InternVL2-8B, Qwen2-VL-7B) achieve end-to-end extraction with strong performance, scoring 95.1/94.1 on SROIE and 88.2/91.2 on CORD.

Our model, PIP-Qwen2-VL-7B, demonstrates the effectiveness of mask pre-training and KV-supervised fine-tuning. It achieves substantial ANLS gains over its base model of +1.4 (77.9 → 79.3) on FUNSD, +2.9 (94.1 → 97.0) on SROIE, and +6.1 (91.2 → 97.3) on CORD. This performance establishes new SOTA records on SROIE (97.0) and CORD (97.3), while remaining highly competitive on FUNSD (79.3). Notably, LayTextLLM's top score on FUNSD relies on idealized OCR input, which limits its practical applicability due to error propagation from real-world OCR systems. In contrast, our end-to-end approach avoids this dependency, offering superior deployment value.

\textbf{Inference Efficiency:}
Autoregressive models like LayTextLLM are constrained by sequential token generation, with per-sample inference times ranging from 0.339s to 1.055s. Our parallel inference paradigm overcomes this bottleneck by generating all target tokens simultaneously. This reduces latency to just 0.028s–0.053s, delivering a 12–20× speedup. Compared to other MLLMs of a similar scale, which require a minimum of 0.198s–0.310s per sample, our method is 5–7× faster, with a maximum latency of 0.028s–0.064s across datasets. These results validate the efficacy of our parallel inference paradigm.

\textbf{Summary:}
Leveraging mask pre-training and KV-supervised fine-tuning, our parallel inference paradigm effectively addresses the performance and efficiency trade-offs of prior methods. Our model achieves SOTA accuracy and significant speed improvements, providing a scalable, high-precision, and low-latency solution for document information extraction.

\begin{table}
\centering
\resizebox{\linewidth}{!}{
\renewcommand{\arraystretch}{1.0}
\begin{tabular}{lcccccccccccccc}
\toprule
 & \multicolumn{2}{c}{\textbf{FUNSD}} & & \multicolumn{2}{c}{\textbf{SROIE}} & & \multicolumn{2}{c}{\textbf{CORD}} & & \multicolumn{2}{c}{\textbf{POIE}} & & \multicolumn{2}{c}{\textbf{WildReceipt}}\\
\cline{2-3} \cline{5-6} \cline{8-9} \cline{11-12} \cline{14-15}
\textbf{Models} & ANLS & Time & & ANLS & Time & & ANLS & Time & & F1 & Time & & F1 & Time \\
\midrule
InternVL2-2B & 60.2 & 0.357 & & 94.4 & 0.332 & & 87.4 & 0.170 & & 77.7 & 0.122 & & 51.3 & 0.184 \\
InternVL2-8B & 61.2 & 0.615 & & 95.1 & 0.541 & & 88.2 & 0.314 & & 77.2 & 0.167 & & 53.8 & 0.347 \\
Qwen2-VL-2B & 75.1 & 0.389 & & 94.5 & 0.458 & & 90.7 & 0.225 & & 87.1 & 0.138 & & 56.2 & 0.182 \\
Qwen2-VL-7B & 77.9 & 0.455 & & 94.1 & 0.532 & & 91.2 & 0.288 & & 86.2 & 0.148 & & 54.9 & 0.218 \\
\hline
\multirow{2}{*}{PIP-InternVL2-2B} & 66.9 & 0.069 & & 92.8 & 0.028 & & 91.0 & 0.018 & & 90.3 & 0.007 & & 64.7 & 0.005 \\
& \textcolor{green}{$\uparrow$6.7} & \textcolor{blue}{5.2×} & & \textcolor{red}{$\downarrow$1.6} & \textcolor{blue}{11.9×} & & \textcolor{green}{$\uparrow$3.6} & \textcolor{blue}{9.4×} & & \textcolor{green}{$\uparrow$12.6} & \textcolor{blue}{17.4×} & & \textcolor{green}{$\uparrow$13.4} & \textcolor{blue}{36.8×}\\
\multirow{2}{*}{PIP-InternVL2-8B} & 72.3 & 0.064 & & 93.4 & 0.034 & & 93.1 & 0.022 & & 93.1 & 0.009 & & 69.0 & 0.010\\
& \textcolor{green}{$\uparrow$11.1} & \textcolor{blue}{9.6×} & & \textcolor{red}{$\downarrow$1.7} & \textcolor{blue}{15.9×} & & \textcolor{green}{$\uparrow$4.9} & \textcolor{blue}{14.3×} & & \textcolor{green}{$\uparrow$15.9} & \textcolor{blue}{18.6×} & & \textcolor{green}{$\uparrow$11.0} & \textcolor{blue}{34.7×} \\
\multirow{2}{*}{PIP-Qwen2-VL-2B} & 75.4 & 0.055 & & 93.4 & 0.049 & & 95.8 & 0.023 & & 92.3 & 0.009 & & 64.8 & 0.006 \\
& \textcolor{green}{$\uparrow$0.3} & \textcolor{blue}{7.1×} & &
\textcolor{red}{$\downarrow$1.1} & \textcolor{blue}{9.4×} & &
\textcolor{green}{$\uparrow$5.1} & \textcolor{blue}{9.8×} & &
\textcolor{green}{$\uparrow$5.2} & \textcolor{blue}{15.3×} & &
\textcolor{green}{$\uparrow$8.6} & \textcolor{blue}{30.3×}\\
\multirow{2}{*}{PIP-Qwen2-VL-7B} & 79.3 & 0.053 & & 97.0 & 0.051 & & 97.3 & 0.028 & & 94.6 & 0.010 & & 71.6 & 0.006 \\
& \textcolor{green}{$\uparrow$1.4} & \textcolor{blue}{8.6×} & & \textcolor{green}{$\uparrow$2.9} & \textcolor{blue}{10.4×} & & \textcolor{green}{$\uparrow$6.1} & \textcolor{blue}{10.3×} & & \textcolor{green}{$\uparrow$8.4} & \textcolor{blue}{14.8×} & & \textcolor{green}{$\uparrow$16.7} & \textcolor{blue}{36.3×} \\
\bottomrule 
\end{tabular}}
\caption{Results of different base models and our PIP-Models for KIE tasks. \textcolor{green}{Green} indicates the performance improvement of our parallel model compared to the base model, \textcolor{red}{red} signifies a decline in performance, and \textcolor{blue}{blue} represents the speedup ratio. }
\label{tab:baseline_results}
\end{table}

\subsubsection{Compared with Base Models}
To demonstrate the generalizability of our paradigm, we conducted experiments across different base models, model sizes, and datasets. The results are presented in Table \ref{tab:baseline_results}.

\textbf{Extraction Performance:}
Our method consistently maintains or enhances the strong extraction capabilities of base models like the Qwen2-VL and InternVL2 series. For example, PIP-Qwen2-VL-7B improves the F1 score on the WildReceipt dataset by a significant 16.7 points over its base model. Performance remains stable or improves across nearly all tasks, with only a minor degradation observed on SROIE.

\textbf{Inference Efficiency:}
As shown in Table \ref{tab:baseline_results}, our PIP-Models achieve over a 5× speedup compared to their autoregressive base models. While standard models generate outputs for each key sequentially, our parallel paradigm generates them simultaneously. Consequently, the acceleration factor scales with the number of keys to be extracted. On key-intensive datasets like WildReceipt, this results in an impressive 36× speedup, achieved without compromising extraction accuracy.

\textbf{Summary:}
In summary, our parallel inference paradigm, combined with its associated training strategies, significantly accelerates inference speed (5–36×) while maintaining or improving extraction accuracy. The approach demonstrates high flexibility and is applicable across various model architectures and sizes.

\begin{table}
\centering
\resizebox{0.8\linewidth}{!}{
\renewcommand{\arraystretch}{1.0}
\begin{tabular}{lccc}
\toprule
\textbf{Model} & \textbf{FUNSD} & \textbf{SROIE} & \textbf{CORD} \\
\midrule
InternVL2-2B & 21.04G & 15.01G & 14.84G \\
InternVL2-8B & 32.89G & 28.45G & 27.75G \\
Qwen2VL-2B & 21.15G & 15.16G & 14.28G \\
Qwen2VL-7B & 32.60G & 28.02G & 27.23G \\
\hline
PIP-InternVL2-2B & 24.39G (+16\%) & 17.83G (+19\%) & 19.09G (+29\%) \\
PIP-InternVL2-8B & 37.17G (+13\%) & 31.99G (+12\%) & 33.56G (+21\%) \\
PIP-Qwen2-VL-2B & 24.58G (+16\%) & 17.91G (+18\%) & 18.63G (+30\%) \\
PIP-Qwen2-VL-7B & 36.71G (+13\%) & 32.52G (+16\%) & 33.22G (+22\%) \\
\bottomrule
\end{tabular}}
\caption{GPU memory usage of different base models and our PIP-Models on FUNSD, SROIE, and CORD datasets. Numbers in parentheses indicate the percentage increase compared to corresponding base models.}
\label{tab:gpu_memory}
\end{table}

\begin{figure}[ht]
\centering 
\includegraphics[width=\linewidth]{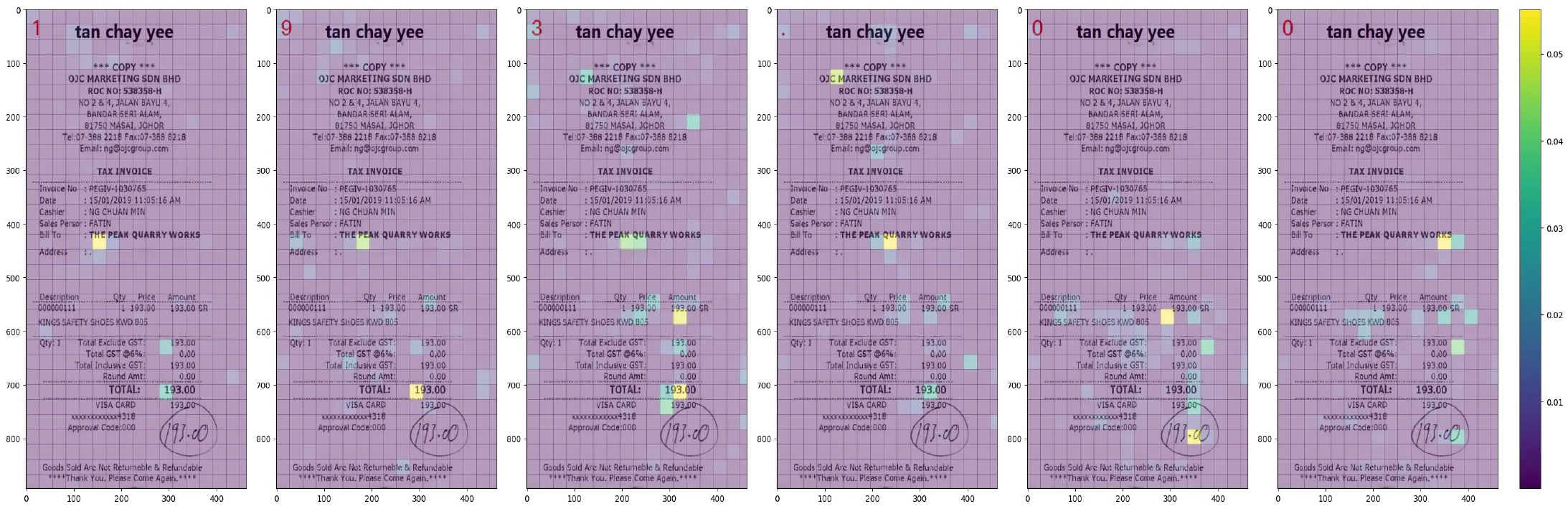}
\caption{The visualization of attention for each token in our PIP-Models when outputting "193.00", demonstrating the model's focus on different regions of the image.}
\label{fig:attention_visualization}
\end{figure}

\subsection{GPU Memory Consumption}
Table~\ref{tab:gpu_memory} compares the GPU memory consumption of baseline models and PIP-Models during inference. The integration of additional “[mask]” tokens to enable parallel output generation increases input length and thus raises memory usage. This overhead remains moderate, with memory usage increasing by at most 30\% over the base models. Nonetheless, the parallel inference enabled by these tokens yields substantial improvements in inference speed (5-36×), leading to higher effective GPU utilization. As shown in Tables~\ref{tab:baseline_results} and~\ref{tab:gpu_memory}, the proposed framework achieves significant efficiency gains with limited memory overhead and negligible impact on extraction performance.

\section{Attention Visualization}
\label{sec:attention_visualization}

To further analyze the parallel inference paradigm for KIE, we visualize the attention maps of output tokens generated by our PIP-Models. As KIE mainly requires extracting information from predefined regions rather than sequential reasoning, output tokens attend to distinct regions of the input image corresponding to different fields. For example, in the SROIE dataset, when extracting the value ``193.00'' for the key ``TOTAL'', each output token attends to relevant image regions, as illustrated in Figure~\ref{fig:attention_visualization}.

While the model does not focus exclusively on the exact answer regions, each token consistently attends to areas pertinent to its target output. For instance, the token predicting ``3'' attends to the region containing the character ``3''. This validates the effectiveness of the parallel inference paradigm for KIE, where output tokens are conditionally independent and require only localized visual context to achieve accurate extraction, supporting efficient parallel generation.

\section{Conclusion}
This paper presents PIP, a simple yet effective parallel inference paradigm for KIE from VrDs. Conventional autoregressive approaches are constrained by sequential token generation, limiting scalability for large-scale document processing. To overcome this, we propose a method that incorporates ``[mask]'' tokens in the input, allowing for the simultaneous generation of all target tokens. We further introduce a tailored mask pre-training scheme and a KV supervised fine-tuning strategy to enhance overall model performance. Our PIP-Models achieve state-of-the-art results on benchmark datasets such as SROIE and CORD. Extensive experiments demonstrate that our approach maintains competitive accuracy while achieving a 5-36× speedup compared to autoregressive baselines, making it well-suited for practical, large-scale deployments. By reconciling efficiency with high accuracy, this work marks a significant advancement in KIE, and provides a scalable solution for real-time document understanding in industry applications.

\section{Acknowledgements}
This work was supported by Ant Group Research Fund, the Joint Funds of the National Natural Science Foundation of China (Grant No.U21B2020) and Science and Technology Cooperation Program of Shanghai Jiao Tong University in Inner Mongolia Autonomous Region——Action Plan of Shanghai Jiao Tong University for "Revitalizing Inner Mongolia through Science and Technology".

\bibliographystyle{IEEEbib}
\bibliography{strings,refs,custom}

\end{document}


%
\maketitle
\section{Implementation Details}
The MLLMs used as baselines in our experiments are Qwen2-VL and InternVL2. Qwen2-VL is available in two versions (2B and 7B), while InternVL2 includes versions of 2B and 8B. During mask pre-training, the masking ratio $\lambda$ was set to 0.5. Both mask pre-training and KV supervised fine-tuning stages used a learning rate of $5 \times 10^{-5}$, with the number of epochs set to 1. Throughout training, the image encoder was frozen, and only the language model are updated. Training processes were conducted using PyTorch 2.4.1 on a cluster of 4 nodes, each equipped with 8$\times$A100 GPUs.

\section{Parameter Analysis}

In this section, we investigate the effect of varying mask ratios $\lambda$ during the mask pre-training phase on the model's performance in downstream information extraction tasks. We conduct mask pre-training on Qwen2-VL-2B with varying mask ratios, and the downstream test dataset used is POIE.

\begin{figure}[ht]
\centering 
\includegraphics[width=\linewidth]{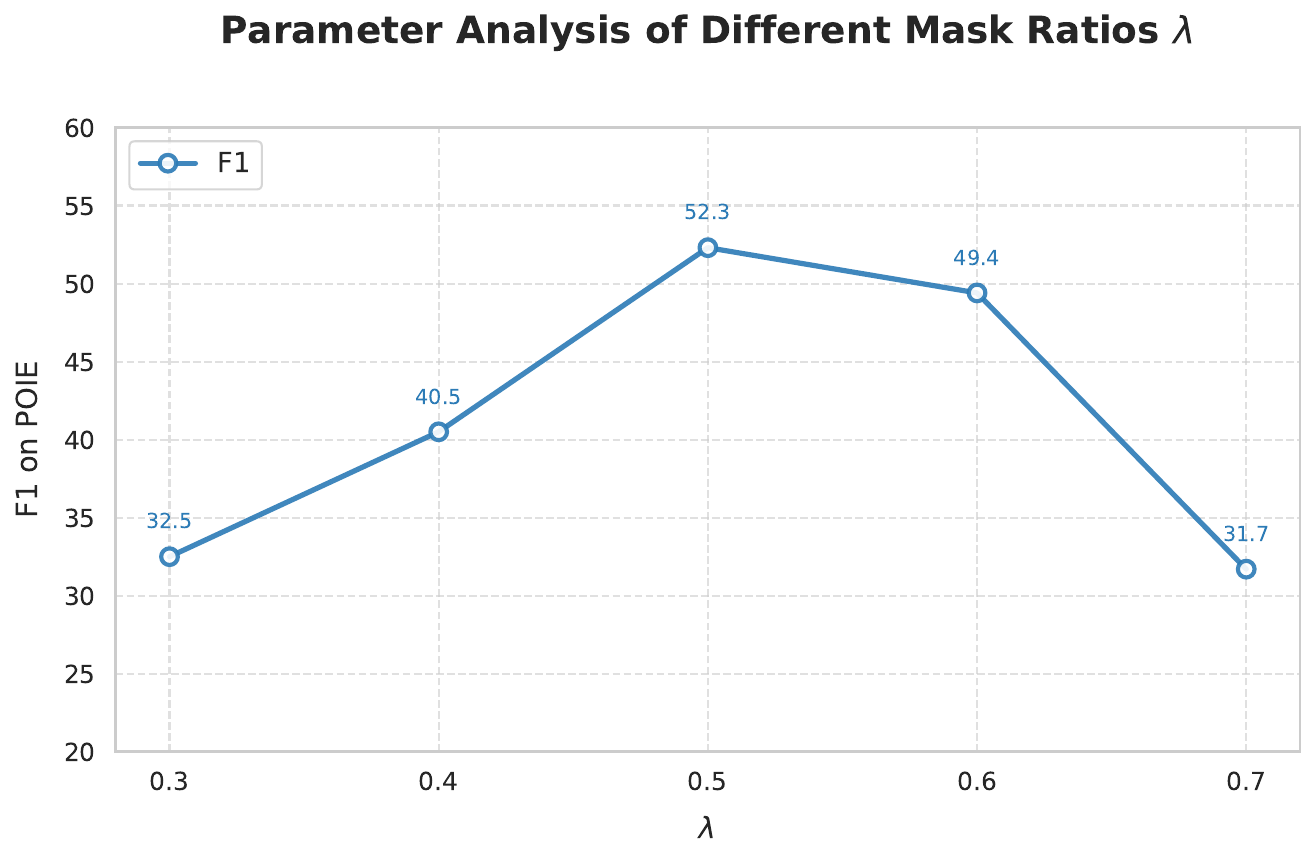}
\caption{Parameter analysis of difference mask ratios $\lambda$ in mask pre-training.}
\label{fig:parameter_analysis}
\end{figure}

As illustrated in Figure \ref{fig:parameter_analysis}, the model achieves optimal performance when the mask ratio $\lambda$ is set to 0.5. For mask ratios below 0.5, the limited number of masked tokens restricts the model's ability to effectively learn the characteristics of masked reasoning. On the other hand, when the mask ratio exceeds 0.5, excessively large masked segments reduce the availability of contextual information in the input data. This hinders the model's ability to accurately reconstruct the original tokens corresponding to the masked positions, leading to suboptimal training outcomes.

\begin{table}[t]
\centering
\resizebox{1\linewidth}{!}{%
\renewcommand{\arraystretch}{1.0}
\begin{tabular}{clcccc}
\toprule
 & \textbf{Models} & \textbf{DocVQA} & & \textbf{VisualMRC} \\
\midrule
\multirow{4}{*}{\textbf{LLMs}} 
& Llama2-7B-chat & $65.0^*$ & & $52.8^*$ \\
& Vicuna-1.5-7B & $67.0^*$ & & $52.1^*$ \\
& LayoutLLM-7B & $74.3^*$ & & $\textbf{55.8}^*$ \\
& LayTextLLM-7B & $77.2^\sim$ & & - \\
\hline
\multirow{1}{*}{\textbf{MLLMs}} 
& LLaVA-1.5-7B & $13.3^*$ & & $35.2^*$ \\
\hline
\multirow{2}{*}{\textbf{Ours}} 
& PIP-InternVL2-8B & 73.4 & & 52.3 \\
& PIP-Qwen2-VL-7B & \textbf{78.7} & & 53.2 \\
\bottomrule
\end{tabular}}
\caption{Results of different models for VQA tasks. * denotes results from \cite{layoutllm}, and $\sim$ indicates results from \cite{laytextllm}. Best results are bolded.}
\label{tab:VQA_results}
\end{table}

\section{Results of VQA Tasks}
Our PIP-Models demonstrate strong performance on the KIE task, substantially improving inference speed while maintaining extraction accuracy. This aligns with our intuition, as we have argued at the beginning of this paper that the characteristics of KIE tasks are well-suited to parallel extraction: different ``[mask]'' tokens only need to attend to distinct image regions and can thus generate outputs concurrently. Subsequent attention visualizations further substantiate this point. It remains an open question whether our PIP-Models can maintain high performance when transferred to other tasks. We therefore aim to investigate the generalizability of the parallel inference paradigm.

We focus on the Visual Question Answering (VQA) task, which not only requires the model to comprehend image content but also demands reasoning to arrive at the final answer. For instance, given an invoice image, a KIE task would involve extracting the specific price of an item, whereas a VQA task might ask whether the price of one item is higher than that of another. In this scenario, the model can no longer directly locate the answer from the image but must instead perform reasoning to derive the result. To evaluate whether our parallel inference paradigm remains effective in this context, we tested our model on two widely-used datasets: DocVQA \cite{docvqa} and VisualMRC \cite{visualmrc}. The experimental results are summarized in Table~\ref{tab:VQA_results}.

As shown in Table~\ref{tab:VQA_results}, among LLMs, Llama2-7B-chat and Vicuna-1.5-7B exhibit moderate performance but struggle with visually intensive document question-answering tasks. In contrast, LayoutLLM and LayTextLLM, which incorporate document layout information, demonstrate improved comprehension capabilities.

Our parallel model outperforms all other autoregressive approaches on DocVQA, and performs slightly below LayoutLLM on VisualMRC. These results indicate that the parallel inference paradigm remains effective even when transferred to other tasks. Furthermore, it is noteworthy that our fine-tuning dataset is specifically constructed for KIE tasks. If parallel extraction were trained on a broader range of VQA datasets, the model's performance could likely be further improved. In future work, we will investigate the performance and generalizability of the parallel inference paradigm across additional tasks to enable its broader application.

\bibliographystyle{IEEEbib}
\bibliography{strings,refs,custom}